\documentclass[11pt]{article}

\usepackage{coling2018}
\usepackage{times}
\usepackage{url}
\usepackage{latexsym}

\usepackage{array}

\usepackage{amsfonts}
\usepackage{amsmath}
\usepackage{subfigure}
\usepackage{graphicx}
\usepackage{CJK}

\title{Source-Critical Reinforcement Learning for Transferring Spoken Language Understanding to a New Language}

\author{
	He Bai$^{1,2}$, 
	Yu Zhou$^{1,2}$,
    Jiajun Zhang$^{1,2}$,
    Liang Zhao$^{3}$,
    Mei-Yuh Hwang$^{3}$ and
	Chengqing Zong$^{1,2,4}$
	\\ 
	$^1$ National Laboratory of Pattern Recognition, Institute of Automation, CAS, Beijing, China\\
	$^2$ University of Chinese Academy of Sciences, Beijing, China\\
	$^3$ Mobvoi AI Lab, Seattle, US\\
	$^4$ CAS Center for Excellence in Brain Science and Intelligence Technology\\
	\{he.bai, yzhou, jjzhang, cqzong\}@nlpr.ia.ac.cn, \{liangzhao, mhwang\}@mobvoi.com
}
\date{}

\begin{document}
\maketitle
\begin{abstract}

To deploy a spoken language understanding (SLU) model to a new language, language transferring is desired to avoid the trouble of acquiring and labeling a new big SLU corpus.
Translating the original SLU corpus into the target language is an attractive strategy. However, SLU corpora consist of  plenty of semantic labels (slots), which general-purpose translators cannot handle well, not to mention additional culture differences. 
This paper focuses on the language transferring task given a tiny in-domain parallel SLU corpus. The in-domain parallel corpus can be used as the first adaptation on the general translator. But more importantly, we show how to use reinforcement learning (RL) to further finetune the adapted translator, where translated sentences with more proper slot tags receive higher rewards. 
We evaluate our approach on Chinese to English language transferring for SLU systems. The experimental results show that the generated English SLU corpus via adaptation and reinforcement learning gives us over 97\% in the slot F1 score and over 84\% accuracy in domain classification. It demonstrates the effectiveness of the proposed language transferring method.  Compared with naive translation, our proposed method improves domain classification accuracy by relatively 22\%, and the slot filling F1 score by relatively more than 71\%.
\end{abstract}
\section{Introduction}
\label{intro}

%
%

\blfootnote{
  %
  %
  %
  %
  %
  \hspace{-0.65cm} 
  This work is licensed under a Creative Commons 
  Attribution 4.0 International License.
  License details:
  \url{http://creativecommons.org/licenses/by/4.0/}
}

Spoken language understanding (SLU) is a key technique in today's conversational systems such as Apple Siri, Amazon Alexa and Microsoft Cortana. To make these conversational systems support multiple languages over different markets, collecting and annotating a large SLU training corpus per language are tedious and costly, and thus hinders the scalability of these systems. It would be greatly helpful if the efforts taken to develop one SLU system could be reused for other languages.

For such a purpose, much work has been reported to explore language transferring of SLU systems or multilingual SLU systems \cite{garcia2012combining,calvo2013exploiting,calvo2016multilingual,jabaian2016unified}. These systems can be grouped into two categories: 
test-on-source-model vs. train-on-target-language. Test-on-source-model is to translate the test sentence in the 2nd language (referred to as L2 from now on) into the first SLU system's language (as L1), and then process it with the L1 SLU system. Train-on-target-language is to translate the L1 training corpus into L2, and then train an SLU system in L2. The train-on-target-language strategy allows tuning and adaptation of the models in the target language directly, and it avoids an overhead of machine translation during real-time execution. The test-on-source-model implies that the final search engine has to deal with L1 while the target answer database might be still in L2. Either the slot value needs to be translated again back to L2, or the database needs to be pre-translated into L1. Hence train-on-target-language is our preferred approach.

Each sample of the SLU training corpus consists of a query and its semantic annotation, for example,``Play $<$song$>$Sorry$<$/song$>$ in $<$album$>$They Don't Know$<$/album$>$''. In the L1 SLU training corpus, both query and its semantic annotation need to be transferred properly to the target language. Literally sending the annotated sentence to general-purpose machine translator or web translator may lose or screw up the annotation information. Two schemes \cite{jabaian2013comparison} have been proposed to solve this problem: transferring source language annotation indirectly through word alignment, vs. adapting the translation model so that it learns how to translate text with slot labels. Each scheme has its own limitations. The first scheme is weak for distant language pairs (e.g. Chinese and English) and the errors are accumulative after translation and word alignment. The second scheme needs extra data for adapting the translation model.  In addition, some previous works \cite{servan2010use,misu2012bootstrapping,jabaian2013comparison} only focused on transferring slot labels but ignored the fact that the slot values should be adjusted properly to the target culture. For example, people in London will probably say ``call a taxi to Tower of London" rather than ``call a taxi to Forbidden City". It's better to make some adaptations on such culture difference for the L2 SLU corpus. In this paper, we will address culture adaptation when we transfer Chinese corpora to  English and we will demonstrate its importance via experimental results. Given that neural machine translation (NMT) is the state-of-the-art translator, NMT model is applied in this paper for language transferring.

The main contribution of this paper is the proposal of using reinforcement learning to improve translation for language transferring. We name it source-critical reinforcement translation (SCRT). We first define
the slot keeping ratio (SKR) as a metric for evaluating the performance of slot transferring. SCRT adapts NMT models via rewarding those translation candidates with higher SKR.
By doing so, we can obtain target language translations which maintain both the semantics and slot information of the SLU labeled sentences in the source language. We evaluate our method on domain classification and slot filling tasks. The results show that our RL method improves the slot F1 score from 93\% to 97\% and domain accuracy from 82\% to 84\%, on top of an already adapted NMT.

\section{Related Work}
Language transferring for SLU systems has been an active research topic in recent years and much work has been reported on both the test-on-source-model and train-on-target-language two strategies. In \newcite{jabaian2013comparison}, different approaches based on test-on-source-model and train-on-target-language strategies are compared. 

The simplest way for the test-on-source-model scheme is to translate L2 target sentences with a web translator into L1, and process these translations with the L1 SLU system. In \newcite{he2013multi} the Microsoft Bing translator is used for this purpose. This approach provides translations of the user input at a very low economic and time cost. Moreover, \newcite{stepanov2013language} have demonstrated that application of language-style and domain adaptation techniques to the ``off-the-shelf and out-of-domain" SMT system could yield improved translation and thus obtain better SLU performance. For adapting MT, \newcite{garcia2014obtaining} translated the L1 SLU training corpus into L2, with multiple web translators to obtain a large parallel dataset, and then trained their own L2-to-L1 SMT model to reduce translation errors. Calvo et al. \shortcite{calvo2013exploiting,calvo2016multilingual} proposed to build graphs of words from different translations and then to parse the translation graph with the SLU model.

The train-on-target-language approach relies on the accurate transferring of semantic annotations from L1 to L2. \newcite{garcia2012combining} translates each segment within one slot separately with a web translator, and then joins these pieces together into a sentence. This method doesn't need additional procedures to transfer slot labels. However, this method misses the context information that is crucial to translation quality. Moreover, it's difficult to determine the ordering of the translated segments. 
\newcite{misu2012bootstrapping} trains an SMT model using conversational in-domain parallel data, and then translates the entire L1 SLU training corpus to L2. However, bilingual in-domain data is scarce and costly, making it difficult to deliver both  quality and low cost. Jabaian et al. \shortcite{jabaian2010investigating,jabaian2013comparison} propose two language transferring schemes. One is transferring source language annotation indirectly through word alignment. The other one is forcing the SMT model to translate the segmentation and slot labels simultaneously. The authors report that the indirect alignment gives the best performance. However, they also point out that distant language pairs suffer severely in word alignment. Finally rather than relying on automatic machine translation, \newcite{stepanov2014development} prefer using human professional translation services. In \newcite{stepanov2017cross}, they extend their work via crowdsourcing for semantic annotation.

\section{Translation Systems}
This section describes all the translation systems conducted in our experiments. Each system has its
own approach of transferring slot labels to the target sentence. They all start from the same 
general-purpose NMT, trained on a huge L1:L2 parallel corpus without SLU annotation. Hence the general translator usually doesn't do well with input containing
slot labels. 

\subsection{Naive Translation}
The naive translation system uses the given well-trained general-purpose translator as it is.
The slot labels in the L1 SLU corpus are stripped off before entering the general translator.
We then align the words between the source sentence and the translated sentence, in order
to add back slot labels into the translated sentence. This is an indirect slot transferring approach.

Some limitations of the naive translation approach come from both translation and alignment. There are plenty of slot values like song names and contact names in the original SLU corpus. Many of these words are out-of-vocabulary (OOV) for the translation model. Although the translation model can handle these OOV words with sub-word modeling techniques \cite{sennrich2015neural}, there are still many slot values remain to be OOV or mistranslated. For example, the Chinese name \begin{CJK*}{UTF8}{gbsn}``白晓霞''\end{CJK*} in Table \ref{table:Inputs-table} is literally translated to ``white sunshine''. Furthermore, wrong translations might also result in wrong alignments, which will yield inaccurate positions for slot labels. 

\begin{table}[t]
\begin{center}
\begin{tabular}{|m{2.2cm}|m{5.8cm}|m{6.3cm}|}
\hline
\textbf{Methods} & \textbf{Source input} & \textbf{Translation result}\\
\hline
Naive\newline Translation&\begin{CJK*}{UTF8}{gbsn}  我想打个电话给白晓霞\end{CJK*}   &I would like to make a call to telephone number of white sunshine\\
\hline
Token-added Translation& \begin{CJK*}{UTF8}{gbsn}我想打个电话给（\,a 白晓霞\,）\end{CJK*}&I would like to make a call to (\,a white sunshine ) 's telephone number please \\ 
\hline
Class-based\newline / SCRT&\begin{CJK*}{UTF8}{gbsn}我想打个电话给\end{CJK*} \$contact\_name & I would love to make a call to \$contact\_name 's number please\\ 
\hline
\end{tabular}
\end{center}
\caption{\label{table:Inputs-table} Translation examples by different translation systems. }
\end{table}

\subsection{Token-added Translation}
To make the translator be aware of slots, we then propose a token-added translation approach. 
This approach uses some special tokens to mark the segmentation boundary for the slot value in the source sentence. These special tokens are common in both the source vocabulary and target vocabulary of the general translator and
their translation is unique and easy to spot. For example, parentheses and double quotes are good candidates as the special tokens.
Enclosing slot values in source sentences by these special tokens can help identify slot boundaries in the translation outputs. In our example in Table \ref{table:Inputs-table}, the special tokens we choose is a pair of parentheses, where the beginning parenthesis is followed by a single character ``a", representing the slot label (in this case contact\_name). We choose a single English letter for Chinese-to-English transfer because we are almost certain that the English letter will be carried to the target side as it is. After translation, the slot values can then be easily identified.

In token-added translation, no additional word alignment process is required. However such approach relies heavily on the NMT general training data where the special tokens (e.g. parentheses or double quotation marks) are kept in both source and target data. For different language pairs, different special tokens might be chosen for the best translation quality. 
Empirically we find that parentheses with a single English character are highly effective for Chinese to English translation. 

\subsection{Class-based Translation}
\label{classMT}
To better translate the slots, we further propose a class-based translation approach. It is to use a class symbol to replace both the slot label and its slot value in the source sentence, for example, ``Play \$song in \$album". In other words, we generalize the source sentence into a pattern sentence.

The class symbol represents the slot label, but without any specific value, as shown in the last row in Table \ref{table:Inputs-table}. Representing each slot segment with a single symbol has a great advantage of avoiding a multi-word segment to be translated into several non-consecutive segments and not enclosed by the correct slot-label pairs. In order to help the general translator understand these new words (class symbols), we need some new parallel sentences with class symbols to adapt the general translator. Then
for each sentence in the L1 SLU corpus, we first transform it to use class symbols and then translate it with the adapted general translator. The translated sentence will contain class symbols as well. We  then replace these class symbols with slot values appropriate to the target culture.

Therefore unlike the first two systems, the class-based translation model is an adapted translator. It requires a small parallel annotated SLU corpus for adaptation.

\subsection{Source-Critical Reinforcement Translation}
The performance of class-based model relies on whether parallel annotated SLU data is enough or not to translate class symbols correctly. In other words, slots will be missed or mistranslated without enough parallel data. However, in-domain SLU parallel data is scarce. It's best if we can take advantage of the existing L1 SLU corpus to solve the slots missing or mistranslated problem of class-based method. Assume the general translator has been adapted into a class-based
translation model, we now propose another adaptation algorithm based on reinforcement learning that
depends on monolingual instead of bilingual SLU corpora.
Hence we name it source-critical reinforcement translation (SCRT).

Before formally introducing SCRT, it is necessary to first define SKR, a metric named slot keeping ratio for evaluating the slot transferring performance of translation models.
For each (Chinese) sentence $c_i$, its SKR is defined as the number of slots in the translated (English) sentence, $e_i$, divided by the number of slots in $c_i$:
%
%
\[\mbox{SKR}(c_i,e_i)=\frac{\sum_s \min(g(c_i,s),g(e_i,s))}{\sum_s g(c_i,s)}\times 100\% \]
The function \(g(c_i,s)\) is used to count the occurrences of slot \(s\) in sentence \(c_i\). Minimum is taken at the numerator to avoid SKR to be over 100\%.

\begin{figure}[tbp] 
\begin{center} 
\includegraphics[width=0.9\textwidth]{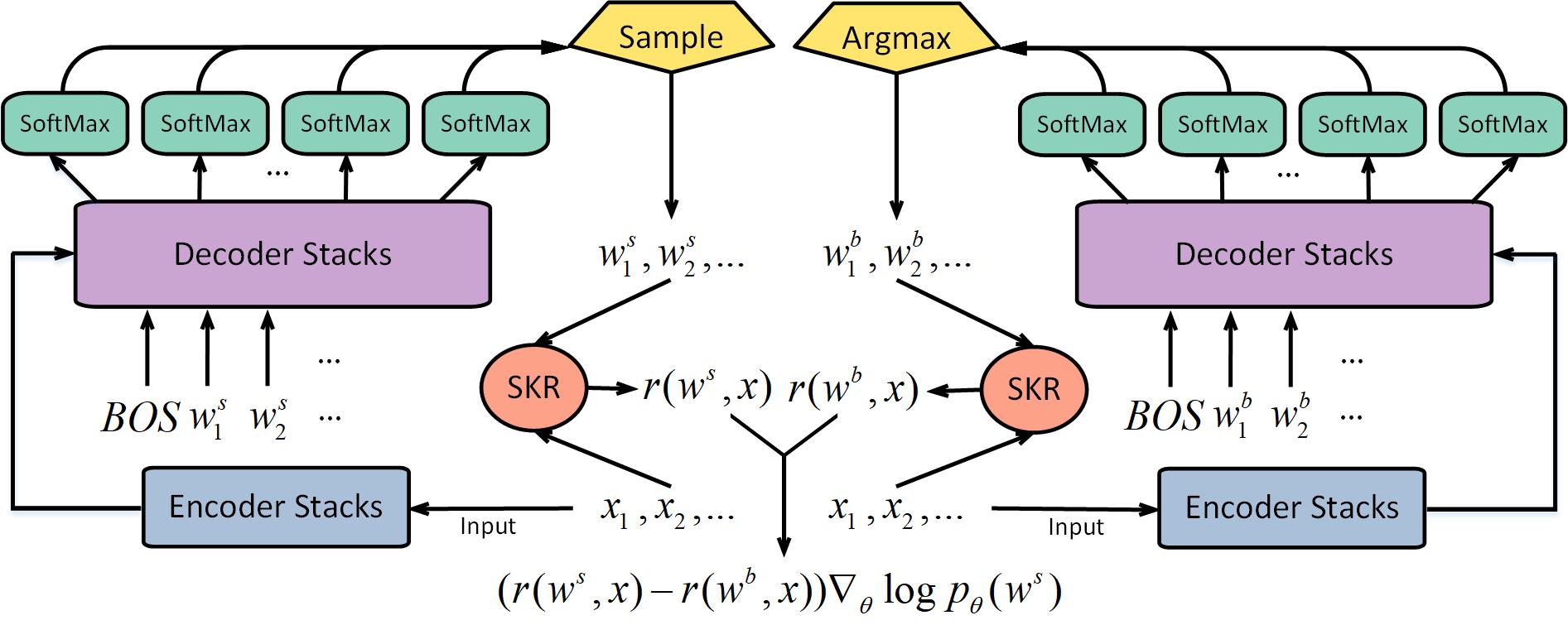} 
\caption{Architecture of Source-Critical Reinforcement Translation. \(x_1, x_2, ...\) is the source input of the NMT model and \((r(w^s,x)-r(w^b,x)) \nabla_\theta\log p_\theta (w^s)\) is the final gradient to optimize with policy gradient method. The encoder and decoder stacks represent inner layers of the Transformer general translator.} 
\label{fig:source-critical} 
\end{center} 
\end{figure}

It's easy to see that a high SKR is a critical condition for generating a high-quality training corpus for L2 SLU models. Our SCRT algorithm directly optimizes the SKR for each source sentence, and learns how to translate specific words according to specific rules. In this task, specific words are those class symbols, and specific rules are the correspondence of each class symbol in the two languages. This is inspired by \newcite{ranzato2015sequence} and \newcite{rennie2016self}. Instead of promoting sequence-level training performance via directly optimizing BLEU scores that require ground truth target sentences, our algorithm focuses on the performance of specific words only, and therefore ground-truth target sentences are not required. The architecture of this model is depicted in Figure \ref{fig:source-critical}. 

Since SCRT is based on REINFORCE algorithm \cite{williams1992simple,zaremba2015reinforcement}, we first describe reinforcement learning from the perspective of sequence generation. 

\subsubsection{REINFORCE}
In order to directly maximize SKR, we can cast our problem in the reinforcement learning framework \cite{sutton1998reinforcement} as in \newcite{ranzato2015sequence}. Our NMT model can be viewed as an \textit{agent} that interacts with an external \textit{environment} (words). The parameters of neural network, \(\theta\), defines a \textit{policy} \(p_\theta\), whose execution results in an \textit{action}. The \textit{action} is the prediction of the next word in the sequence at each time step. After taking an action, our agent will update its internal \textit{state} (attention weights). Once our agent generates the end-of-sequence (EOS) token, the agent will observe a \textit{reward}. We can choose different reward functions for various purposes. Here, we use the SKR value as the reward \(r\).

During training, the agent chooses an action according to the current policy and observes a reward only when the end token is generated. This reward is computed according to the generated sentence and the input sentence. 
The goal of training is to maximize the expected reward, and therefore 
the loss function (negative of reward) for the encoder-decoder given an input source sentence, $x=x_1, x_2, ...$, is defined as:
\begin{equation}\label{equ:expectation}
\begin{split}
L(\theta)&=-\sum_{w^g_1,...,w^g_T}p_\theta(w^g_1,...,w^g_T|x)\,r(w^g_1,...,w^g_T, x)\\
&=-\mathbb{E}_{w^g\sim p(\theta)}r(w^g,x)
\end{split}
\end{equation}
where \(w^g=(w_1^g,...,w_T^g)\) is the generated sentence of length $T$. Notice the length of $x$ is not necessarily $T$. The notion $w^g \sim p(\theta)$ means
the generated sentence follows the distribution of the
decoder softmax output $p(\theta)$.

The gradient of \(L(\theta)\) can be computed as follow:
\begin{equation}\label{equ:deviation}
\nabla_\theta L(\theta) = -\mathbb{E}_{w^g\sim p(\theta)}r(w^g, x) \nabla_\theta\log p_\theta (w^g)
\end{equation}
We approximate this expectation with Monte Carlo based sampling and a baseline \(b\) is used to reduce the variance in the Monte Carlo estimator of the gradient, following \newcite{weaver2001optimal}:
\begin{equation}\label{equ:baseline}
\begin{split}
\nabla_\theta L(\theta) &= -\mathbb{E}_{w^s\sim p(\theta)}(r(w^s,x)-b) \nabla_\theta\log p_\theta (w^s)\\ 
& \approx -(r(w^s,x)-b) \nabla_\theta\log p_\theta (w^s)
\end{split}
\end{equation}
where $w^s$ is the sampled sentence. We use a single sample to approximate the expectation in the above formula. Using the chain rule of gradient computation, we have:
\begin{equation}\label{equ:chainrule}
\nabla_\theta L(\theta) =\sum_t\frac{\partial L_\theta}{\partial y_t}\frac{\partial y_t}{\partial \theta}
\end{equation}
where \(y_t\) is the input to the decoder softmax function at time stamp $t$. Following Zaremba's prior work, the gradient of loss \(L(\theta)\) with respect to \(y_t\) is given by:
\begin{equation}\label{equ:gradient}
\frac{\partial L_\theta}{\partial y_t}=(r(w^s, x)-b)(p_{\theta}(w_{t}|h_t)-1(w_{t}^s))
\end{equation}
where \(h_t\) is the input vector to the fully-connected layer before softmax, and \(1(w_{t}^{s})\) is the one hot vector for \(w_{t}^{s} \). Notice $y_t$,
$p_\theta(w_t|h_t)$, $1(w_t^s)$ are all vectors of length V, where V is the 
number of output units from the decoder.

\subsubsection{Self-Critical and Source-Critical}
The baseline reward $b$ is obtained by the current model using the inference algorithm at test time \cite{rennie2016self}. This is called
 self-critical training. In this work, we choose greedy decoding at test time. That is,
the most likely word from the decoder output $p(\theta)$ at each time stamp is selected
as the baseline output, as shown on the right-hand side of Figure \ref{fig:source-critical}. The greedy translation sentence is denoted as $w^b$.

The above gradient can now be written as:
\begin{equation}\label{equ:final}
\frac{\partial L_\theta}{\partial y_t}=(r(w^s, x)- r(w^b, x))(p_{\theta}(w_{t}|h_t)-1(w_{t}^s))
\end{equation}
Since the reward is based on SKR exclusively, no ground truth target sentence is needed. We call this source-critical learning. 

\subsubsection{Optimization}
Equation (6) says that the sampled sentence \(w^s\) acts like a surrogate target for our output distribution, \(p_{\theta}(w_{t}|h_t)\), at time \(t\). Once the sampled sentence achieves a higher SKR than greedy decoding, a positive reward is used for parameter updating; otherwise, a negative reward is used. 

The SKR criterion itself does not impose constraints on translation quality. Therefore it needs a well-trained general-purpose NMT system as the initial model. 
In this paper, we initialize the SCRT model with parameters from the adapted class-based model described in Section \ref{classMT}, which also
offers good embedding features for the class symbols. Otherwise,
 the model would be difficult to converge, as the sampling space or action space is enormous. 

%
%

After initialization, we then use policy gradient methods to find parameters that lead to a large expected reward. As our SCRT only optimizes the SKR score with monolingual data, the curriculum learning \cite{bengio2009curriculum} strategy is employed in which we begin with the class-based model, and gradually increase the number of SCRT training steps. 

\section{Experiments}
\subsection{Experimental Setup}
We conduct experiments on Chinese to English language transferring task and evaluate the quality of the translated corpus via domain classification and slot filling tasks.

We annotated 3000 (over time this should be a very big number) Chinese dialogue sentences as the L1 SLU corpus and 1500 English dialogue sentences as the test set for the target language. Our goal is to transfer these 3000 Chinese sentences to English, and then train an English SLU model to handle these 1500 English test sentences. 

Next we annotated additional 1500 pairs of parallel sentences for adapting the general translator into the class-based MT. This small parallel SLU corpus is called the adaptation corpus in this section. Finally, the Chinese side of the 3000 sentences of the L1 SLU corpus, together with the Chinese side of the 1500 sentences of the adaptation corpus is used in SCRT training, on top of the adapted class-based MT.

\subsubsection{SLU model}
The SLU annotated data  are collected from communication, navigation and music three domains. It includes seven slots: contact name, contact type, address type, song name, album, feature and artist. As ``others'' domain is necessary for domain classification in practical applications, we  collect additional 40,000 sentences from other scenarios as the fourth domain. For domain classification models, we use the SVM linear classifier with public toolkit LIBLINEAR \cite{fan2008liblinear}. Stop words are removed from the training corpus. Simple N-gram features with a cut-off of 10 are used in domain classification. For slot filling experiments, the CRF++ toolkit \cite{kudo2005crf++} is used. The slot labeling follows the IOB format. 

\subsubsection{The general translator}
We choose Transformer \cite{vaswani2017attention} as the general machine translation model. And for naive translation, we use Fast-Aligner (FA-IBM 2) \cite{dyer2013simple} to locate the slot labels for the translated queries. The architecture of our NMT model is the same as \newcite{vaswani2017attention}.

Our training corpus is from AI Challenger: The English-Chinese Machine Translation track\footnote{https://challenger.ai/datasets/translation}. This competition provides over 10 million parallel English-Chinese sentences which were collected from English learning websites and movie subtitles. In our experiments, we extract nearly 8 million pairs of sentences from this corpus to train the baseline machine translation model. After sub-word \cite{sennrich2015neural} preprocessing, the source vocabulary size is 83,000 and the target part is 78,000. Our training batch size is 3,072 and we train the baseline model for a total of 300,000 steps with Adam optimizer \cite{kingma2014adam} on two GPUs. All decodings are conducted with a beam size of 4, and the top one translation is taken as the final translation output.

\subsubsection{Culture adaptation}
In our experiments, we select the culture-dependent slots such as contact name, song name, album and artist for target culture adaptation. We first collect thousands of English names, artists, song names and albums to build a database. The slot values in the test data are removed from the database. In the translated target queries, the corresponding slots are filled by randomly selected slot values from the database. The same random seed is used for all approaches.
 
\subsection{Results}
\subsubsection{Without parallel adaptation corpus}
We first compare the performance without the parallel adaptation data. Table \ref{table:results_without_adaption} shows that the performance of naive translation method and token-added translation method. Note that the class-based model and the SCRT trained model are not included in this table, because the class-based translator needs the adaptation data to learn the translation of class symbols, and SCRT training is built on top of the class-based model.

In Table \ref{table:results_without_adaption}, compared with naive translation method, the token-added approach achieves a higher SKR score. However, keeping more slots doesn't yield good performance in the slot filling task. More mistranslations may happen in the token-added approach due to the fact that the added special tokens actually change the context of the original query and introduce noisy information.  
On the other hand, the token-added approach has a slightly better domain accuracy than the naive translation. This is probably because slot labels carry a certain
degree of domain information. 

In Table \ref{table:results_without_adaption}, the contribution of cultural adaptation is obvious. After filling the corresponding slots with culture appropriate slot values, substantial improvements are observed in this table. Adding culture adaptation gives naive translation method more than 50\% relative improvement over slot filling F1 score and domain classification accuracy. The similar trend also holds for token-added translation.
\begin{table}[t]
\centering
\begin{tabular}{|c|c|c|c|c|}
\hline
\textbf{Trans.} &Culture A.& SKR&Slot\_F1&Dom\_Acc\\
\hline
Naive & No & 57.13\% &45.39&45.87\%\\
\hline
Naive & Yes & 57.13\% &69.78&78.93\%\\
\hline
TA& No & 60.82\% &24.87&55.07\%\\
\hline
TA&Yes & 60.82\% &34.14&81.4\%\\
\hline
\end{tabular}
\caption{\label{table:results_without_adaption} Slot F1 scores and domain accuracy, with naive translation vs. token-added (TA) translation using an unadapted general translator. The second column indicates whether culture adaptation is applied. }
\end{table}

\begin{table}[t]
\centering
\begin{tabular}{|c|c|c|c|c|}
\hline
\textbf{Trans.} & Culture A. & SKR&Slot\_F1&Dom\_Acc\\
\hline
Naive & No &57.50\% &56.79&68.73\%\\
\hline
Naive & Yes &57.50\% &70.79&81.4\%\\
\hline
TA & No &\textbf{98.55}\% &90.20&66.86\%\\
\hline
TA & Yes &\textbf{98.55}\% &91.91&82.87\%\\
\hline
Class-based& Yes & 97.03\% &93.04&82.12\%\\
\hline
+SCRT&  Yes & 98.08\% &\textbf{97.19}&\textbf{84.2}\%\\
\hline
\end{tabular}
\caption{\label{table:results_adaptation} Results after adapting all translators with the additional 1200 parallel SLU sentences. } 
\end{table}

\begin{table}[t]
\centering
\begin{tabular}{|c|c|c|c|c|}
\hline
\textbf{Trans.} & Culture A. & SKR&Slot\_F1&Dom\_Acc\\
\hline
Naive & No &57.2\% &50.15&76.73\%\\
\hline
Naive & Yes &57.2\% &70.13&\textbf{81.93}\%\\
\hline
TA & No &88.22\% &84.96&59.8\%\\
\hline
TA & Yes &88.22\% &85.21&78.53\%\\
\hline
Class-based& Yes & 85.07\% &83.94&76.8\%\\
\hline
+SCRT&  Yes & \textbf{88.6}\% &\textbf{91.07}&81.46\%\\
\hline
\end{tabular}
\caption{\label{table:results_adaptation_90} Results after adapting all translators with the additional 90 parallel SLU sentences. } 
\end{table}

\subsubsection{With parallel adaptation corpus}
In this section, we first compare different methods that use the adaptation corpus that consists of 1500 in-domain annotated sentences in communication, navigation and music domains. In the adaptation corpus, 1200 sentences are randomly selected for training and the rest as the validation data to terminate training. Although 1200 sentences are tiny for model training, we also conduct the same experiments with a much small amount of data, 90 parallel sentence pairs which are also randomly selected from the 1500 in-domain annotated sentences, to test the performance of different methods under different condition.

In Table \ref{table:results_adaptation}, comparing the naive translation with the token-added translation, we can find that the token-added method benefits more from adapting general purpose translator with in-domain parallel data. The SKR of naive translation barely increased, even using the adapted translator. It is obvious that the main limitation of naive translation is alignment rather than translation. The class-based model achieves 97.03\% SKR after the convergence of supervised training with parallel adaptation data. Finally, SCRT training with 4500 monolingual annotated data is further applied to the adapted class-based translator. The SKR is increased to 98.08\%. The F1 score of slot filling and accuracy of domain classification also jump significantly, which undoubtedly proves that our SCRT can generate better SLU training corpus for other languages, with the aid of monolingual annotated data exclusively. 

In Table \ref{table:results_adaptation_90}, the SCRT achieves the highest F1 score of slot filling and the slot transferring or SKR still benefits from monolingual data with SCRT incremental training. This result shows that our proposed method is very competitive even with an extremely small amount of parallel in-domain data. But the naive translation method with culture adaptation show a better performance on domain classification. This is because, with an extremely small amount of parallel training data, it is hard to achieve a high quality of translation for SCRT and class-based method, which is crucial for domain classification. Besides, the naive translation always benefits more from the culture adaptation in domain classification, as those mis-aligned segments will be replaced with correct slots, which makes such sentences  more distinguish from the other.

\subsubsection{Analysis of SCRT}
\begin{figure}[tbp] 
\begin{center} 
\includegraphics[width=0.6\textwidth]{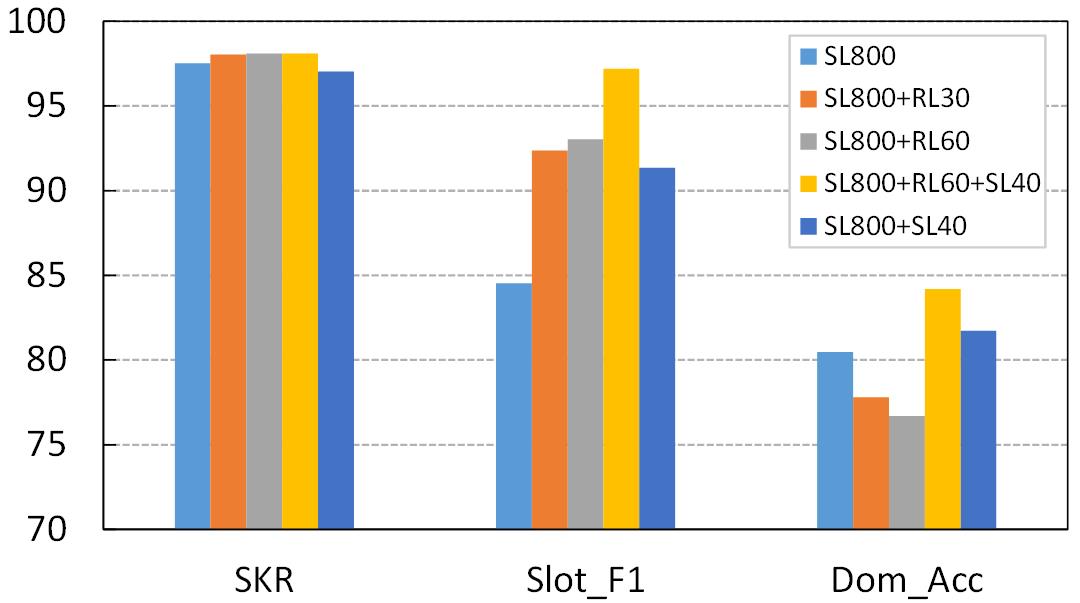} 
\caption{Performance on SKR, slot filling and domain classification, using 
translation models trained at different steps of supervised learning (SL) and reinforcement learning (RL).} 
\label{fig:bar-scores} 
\end{center} 
\end{figure}

In order to evaluate how much the proposed SCRT contributes to the class-based model, we analyze the systems at different iterations of training with 1200 in-domain parallel sentences.
In the experiments, we find that the class-based model begins to converge after 800 steps with a batch size of 32 when adapted on the 1200 in-domain parallel sentence pairs. So this model at step 800 is chosen as a baseline and is labeled as SL800 in Figure \ref{fig:bar-scores}. 
Based on such a baseline, we use the 4500 sentences of monolingual data to conduct 30 steps (SL800+RL30) and 60 steps (SL800+RL60) SCRT RL training. As shown in Figure 2, the SKR increases nearly one percent and the F1 score increases by over 7 percents for both models. However, we note that the accuracy of domain classification drops. This is because SKR imposes no constraints for general machine translation quality that is crucial for domain classification. 
The translation quality can be improved by additional steps of supervised training using in-domain parallel training data. For model SL800+RL60+SL40, another 40 steps of supervised training are conducted on top of SL800+RL60. 

As we can see from Figure 2, SL800+RL60+SL40 achieves the highest scores on both slot filling and domain classification among all these models. The last model, SL800+SL40, is used for comparison with SL800+RL60+SL40. Although these two models are trained with the same steps of supervised learning, the model with SCRT achieves higher scores on all metrics. This result indicates the potential contribution of the monolingual data and the effectiveness of our proposed SCRT algorithm for language transferring. Furthermore, comparing SL800 and SL800+SL40, we can find that although the accuracy and F1 score goes up, the SKR goes down after additional training with parallel data. This is the shortcoming of maximum-likelihood estimation objective function for our task: treating all words equally without emphasizing important words like slot labels. 

\begin{figure}
  \centering
  \subfigure[Class-based]{\includegraphics[width=0.45\textwidth]{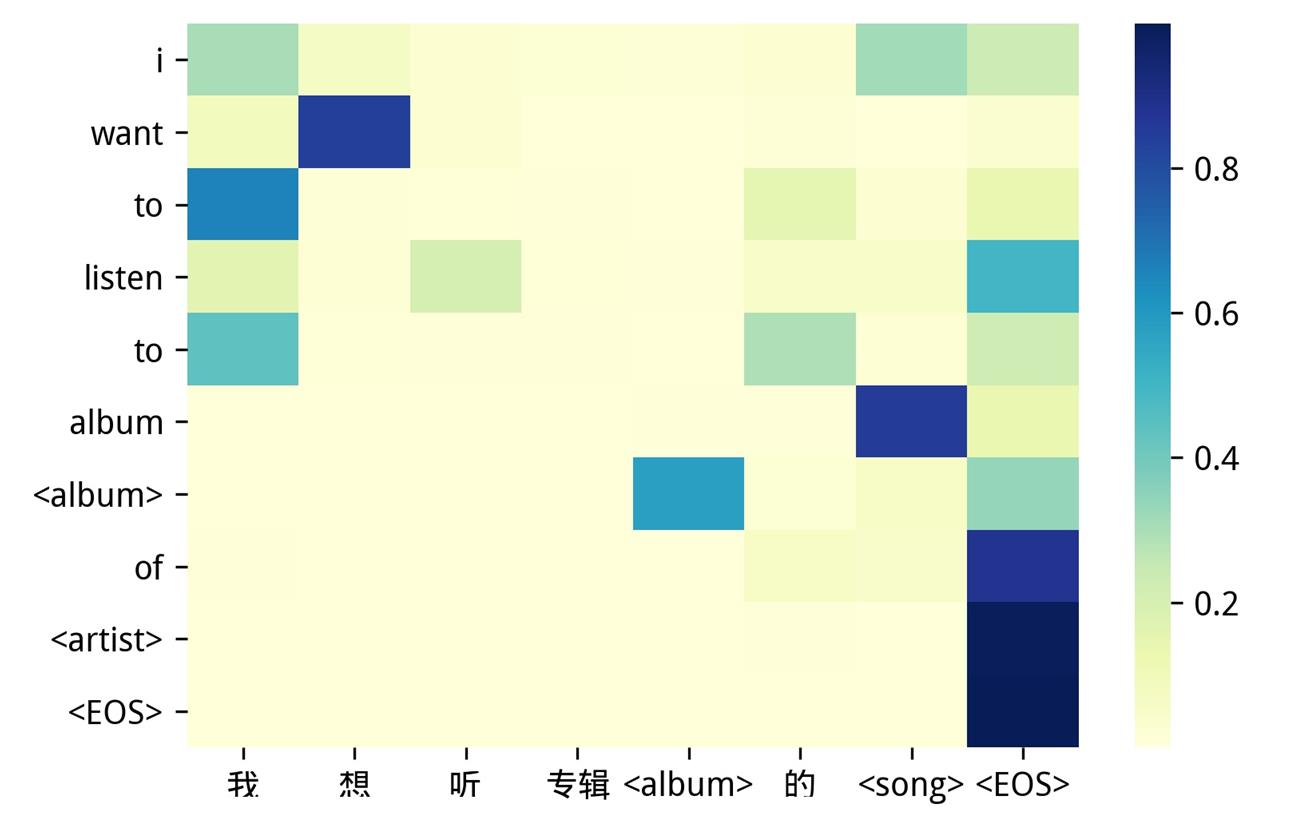}\label{fig:sl heatmap}}
  \subfigure[SCRT]{\includegraphics[width=0.45\textwidth]{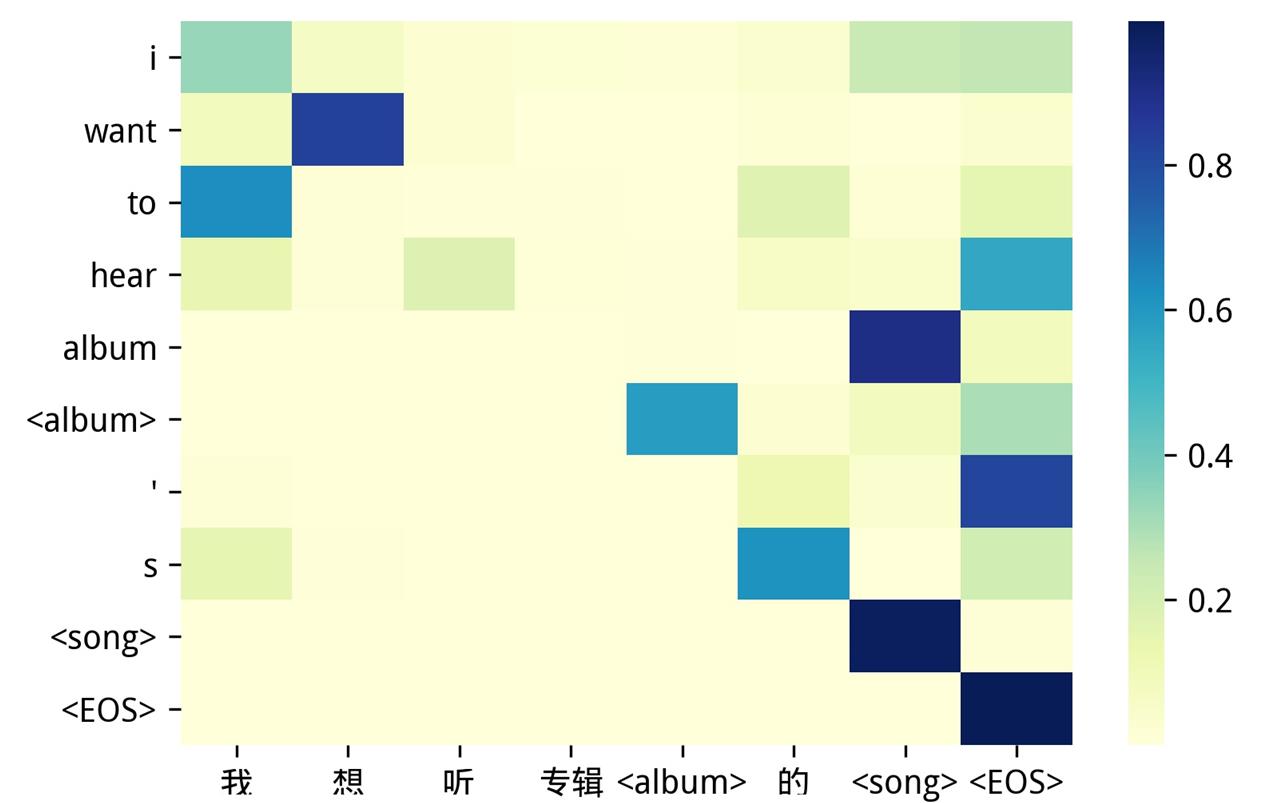}\label{fig:rl heatmap}}
  \caption{Two heat maps of attention vectors between encoder and decoder in the translation model. The x-axis and y-axis of each plot correspond to the words in the source sentence (Chinese) and the generated translation (English), respectively.}
  \label{fig:heatmap}
\end{figure}

The proposed approach SCRT provides an intuitive way to guide the parameter updating for slots transferring with monolingual data. We visualize the attention weights between encoder and decoder of the translation model in Figure \ref{fig:heatmap}. The left plot corresponds to the model SL800 mentioned above and the right one is the model SL800+RL60. 
From the heat maps, we can see which positions in the source sentence were considered more important when generating the target word. Figure \ref{fig:sl heatmap} shows that class-based model mistranslated the slot \$song into \$artist, and was corrected after additional SCRT training in Figure \ref{fig:rl heatmap}. 

\section{Conclusions}
Our work is motivated by the practical demand in language transferring for SLU systems: the lack of large annotated in-domain parallel data, and the requirement of high-quality SLU corpus in the target language. To address this problem, we applied an adapted Neural Machine Translator to translate the SLU corpora to other languages. A small in-domain parallel data is used to adapt the general purpose NMT firstly. Based on the adapted NMT, we proposed a reinforcement learning approach with a source-critical mechanism to do further adaptation using monolingual data exclusively. Our proposed method optimizes the slot keeping ratio directly and adapts slot values accordingly based on the target culture. 
The experiments showed that comparing with naive translation, the proposed method could improve domain classification accuracy by relatively 22\%, and the slot filling F1 score by more than 71\%.

\section*{Acknowledgements}

The research work descried in this paper has been supported by the National Key Research and Development Program of China under Grant No. 2017YFC0822505 and  the Natural Science Foundation of China under Grant No. 61673380.


\bibliographystyle{acl}
\bibliography{coling2018}
\end{document}